%% file: paper.tex
\documentclass{article}
\pdfoutput=1
\usepackage[preprint]{neurips_2021}
\usepackage{algorithm}
\usepackage{caption}
\usepackage{subcaption}
\usepackage[noend]{algpseudocode}
\usepackage{amsfonts}
\usepackage{amsmath}
\usepackage{url}
\usepackage[pdftex]{graphicx}
\usepackage{makecell}
\usepackage{siunitx}

\def\Section {\S}

\newcommand{\squishlist}{
 \begin{list}{$\bullet$}
  { \setlength{\itemsep}{0pt}
     \setlength{\parsep}{3pt}
     \setlength{\topsep}{3pt}
     \setlength{\partopsep}{0pt}
     \setlength{\leftmargin}{1.5em}
     \setlength{\labelwidth}{1em}
     \setlength{\labelsep}{0.5em} } }

\newcommand{\squishlisttwo}{
 \begin{list}{$\bullet$}
  { \setlength{\itemsep}{0pt}
     \setlength{\parsep}{0pt}
    \setlength{\topsep}{0pt}
    \setlength{\partopsep}{0pt}
    \setlength{\leftmargin}{2em}
    \setlength{\labelwidth}{1.5em}
    \setlength{\labelsep}{0.5em} } }

\newcommand{\squishend}{
  \end{list}}
  
\title{A Moment in the Sun: \\Solar Nowcasting from Multispectral Satellite Data using Self-Supervised Learning}
\author{Akansha Singh Bansal\\
UMass Amherst\\
\texttt{akanshasingh@umass.edu}
\And
Trapit Bansal \\
UMass Amherst\\
\texttt{tbansal@cs.umass.edu}
\And
David Irwin\\
UMass Amherst\\
\texttt{deirwin@umass.edu}}
    
\begin{document}


\maketitle

\begin{abstract}
\input{abstract}

\end{abstract}





\section{Introduction}
\label{sec:introduction}
\input{introduction}

\section{Background}
\label{sec:background}
\input{background}
\section{Solar Nowcasting Model Design}
\label{sec:design}
\input{design}

\subsection{Forecasting Solar Output}
\input{model_solar}

\section{Implementation}
\label{sec:implementation}
\input{implementation}

\section{Evaluation}
\label{sec:evaluation}

\input{evaluation}

\section{Related Work}
\label{sec:related}
\input{related}

\section{Conclusion}
\label{sec:conclusion}
\input{conclusion}
\bibliographystyle{ACM-Reference-Format}
\bibliography{paper,example_paper} 

\end{document}

%% file: abstract.tex
Solar energy is now the cheapest form of electricity in history.  Unfortunately, significantly increasing the grid's fraction of solar energy remains challenging due to its variability, which makes balancing electricity's supply and demand more difficult.  While thermal generators' ramp rate---the maximum rate that they can change their output---is finite, solar's ramp rate is essentially infinite.  Thus, accurate near-term solar forecasting, or \emph{nowcasting}, is important to provide advance warning to adjust thermal generator output in response to solar variations to ensure a balanced supply and demand.  To address the problem, this paper develops a general model for solar nowcasting from abundant and readily available multispectral satellite data using self-supervised learning. 
Specifically, we develop deep auto-regressive models using convolutional neural networks (CNN) and long short-term memory networks (LSTM) that are globally trained across multiple locations to predict raw future observations of the spatio-temporal data collected by the recently launched GOES-R series of satellites. 
Our model estimates a location's future solar irradiance based on satellite observations, which we feed to a regression model trained on smaller site-specific solar data to provide near-term solar photovoltaic (PV) forecasts that account for site-specific characteristics.
We evaluate our approach for different coverage areas and forecast horizons across 25 solar sites and show that our approach yields errors close to that of a model using ground-truth observations.

%% file: introduction.tex
Solar is now the cheapest form of electricity in history.  As a result, the U.S. Energy Information Administration (EIA) projects that the share of renewable energy from solar and wind in the grid will double to almost 42\% by 2050 with solar poised to account for nearly 80\% of this increase~ \citep{eia2021}.  This dramatic increase in solar generation is also necessary to mitigate the negative environmental and economic consequences of climate change due to carbon emissions from thermal generators.  Unfortunately, significantly increasing the grid's fraction of solar energy, e.g., beyond 50\%, remains challenging due to its variability, which makes balancing electricity's supply and demand more difficult.  While thermal generators' ramp rate---the maximum rate they can change their output---is finite, solar energy's ramp rate is essentially infinite.  As a result, to maintain the grid's frequency and voltage within a narrow range, utilities will require accurate near-term forecasts of solar generation that provide advance warning to compensate for any changes, either by adjusting thermal generator output or using demand response.


To address the problem, this paper develops a general model for near-term solar forecasting, or \textit{nowcasting}, from multi-spectral satellite data using deep learning. 
Specifically, we make use of real-time data obtained from the GOES-R series of satellites, in particular GOES-16. This satellite observes the continental U.S.A in sixteen spectral bands of light, generating rich spatio-temporal data at an unprecedented temporal and spatial resolution: every five minutes for every 0.5-2km$^2$ area.
This data presents an untapped opportunity for solar nowcasting at the continental scale using data-intensive deep learning techniques, which have as yet been limited to sky-imagery data \citep{zhang2018deep,zhao20193d,siddiqui2019deep,paletta2020convolutional} collected using ``site-specific'' specialized hardware, which inhibits their applicability.
In particular, we utilize the intuition that the first three spectral bands of light, corresponding to the visible region, capture information about the solar irradiance and cloud cover at the observed area.
Thus, we propose to learn a global deep autoregressive model, i.e., a model that predicts the next observation in a sequence, directly from the raw satellite observations to capture the statistical patterns that are indicative of future solar irradiance.



Our intuition is similar to that of prior work on solar nowcasting using cloud motion vectors.  Clouds are the primary reason solar sites' output drops from its clear sky potential, which is largely deterministic based on the ambient temperature, time-of-day, day-of-year, and location.  While cloud movements are a function of complex non-linear atmospheric dynamics over long time periods, their movements are more predictable over short periods \citep{lorenz2004short,cros2014}.  Thus, solar forecasting over long periods, e.g., a few hours to days, requires Numerical Weather Prediction (NWP) models \citep{richardson1922weather} that use physical atmospheric models to account for non-linear dynamics. In contrast, solar nowcasting over short time periods can be easier to model due to the larger influence of more recent changes.  In particular, prior work on solar nowcasting has focused on programmatically identifying clouds in satellite or sky images to determine their size, direction, and velocity~\citep{lorenz2004short,cros2014}. Solar nowcasting can use such cloud motion vectors to forecast solar output based on the direction and velocity of clouds.  Our approach has a similar intuition, but instead of directly identifying cloud motion vectors for which there is no training data, we train a deep learning model that takes as input historical spatio-temporal satellite observations of a region to infer how spectral satellite data is changing over time and space.  Changes in this spectral data implicitly capture cloud movements, as clouds reflect more light, which satellites capture in the spectral data. 

Based on our intuition above, we develop self-supervised deep learning models using convolutional neural networks (CNNs) and long short-term memory networks (LSTM) to forecast the next satellite spectral values at the solar sites of interest. 
These models require historical spectral satellite data over a region surrounding the site as input. We analyze and quantify model accuracy based on both the amount of temporal data, i.e., how far in the past, and the size of the region, i.e., how large of a region, used as input for forecasting 15 minutes in the future.  As we show, the more distant the forecast horizon, the larger the historical data and spatial region required, and the lower the accuracy. However, there are rapidly diminishing returns in accuracy improvement, and significant increases in training time and resources, once the historical data and spatial region reach a certain size.  We then apply a simpler regression model to infer a specific site's solar output from its spectral forecast data obtained from the self-supervised CNN-LSTM model.  We compare our solar nowcasting models with both the accuracy of this model, which infers solar output based on current conditions, as well as a persistence model that assumes that the future power remains unchanged over the forecasting horizon, which also serves as the baseline for comparisons. 

Importantly, we condition our analysis above based on the magnitude and frequency of changes in solar output at a given location.  Put simply, if a location, such as San Diego, California, U.S.A., is rarely cloudy, then a simple persistence approach that predicts near-term solar output never changes will be highly accurate.  Even in highly variable climates, solar output often does not change much over short time periods of 5-30 minutes, which makes simple persistence models appear highly effective.  However, accurately forecasting ``big'' changes in solar output is most important for grid operations, as these are the changes that require an active response.   As a result, we specifically focus on the accuracy of forecasting ``big'' near-term changes in solar power.   As we show, the larger the change in solar output, the larger the improvement in accuracy between our deep learning approach compared with others. 

Our hypothesis is that solar nowcasting using deep learning models trained on multispectral satellite data is both more general and more accurate than prior solar nowcasting approaches, especially at forecasting large changes in solar output.  In evaluating our hypothesis, we make the following contributions. 

\noindent {\bf Satellite Data Compilation.}  We compile a large-scale dataset for 25 solar sites that includes their average solar power generation, ambient temperature, and satellite data across 16 spectral channels for their surrounding region (up to 10km away) every 5 minutes for a year-long period.  We use this dataset to train and test our deep learning models, and plan to publicly release it.



\noindent {\bf Self-supervised Models on Satellite Data.} We develop self-supervised deep learning models trained on spectral satellite observations that use convolutional neural networks (CNNs) and long short-term memory networks (LSTM) to forecast the future observations. These models utilize the spatio-temporal observations across all 25 sites (analyzed in this work) for large-scale training. We analyze the importance of both spatial and temporal components, and also compare with other simpler machine learning methods for such self-supervised modeling.

\noindent {\bf Deep Solar Nowcasting Models.} 
We demonstrate the utility of self-supervised models for solar nowcasting, which depends on factors like solar irradiance and cloud cover. These models use the forecasted spectral data from the self-supervised model as input to a separate site-specific regression model that predicts a specific site's solar output at 15 minutes in the future from current spectral satellite data. This regression incorporates the effect of physical site characteristics, such as module area, tilt, orientation, and tree cover, on solar output.

\noindent {\bf Implementation and Evaluation.}  We implement our models above in python using Tensorflow \citep{abadi2016tensorflow}, and  train them on a GPU cluster.  Given the size of the datasets and complexity of our models, training each model requires $\sim$86 GPU-hours.  We evaluate our approach for different coverage areas and forecast horizons across the 25 solar sites, and show that our approach yields errors close to that of a model that uses ground truth observations.  We also show that our deep learning models are much more accurate at identifying ``big'' changes in near-term solar output.

%% file: background.tex
\begin{figure}[t]
\begin{minipage}[t]{0.48\linewidth}
\centering
\includegraphics[width=\textwidth]{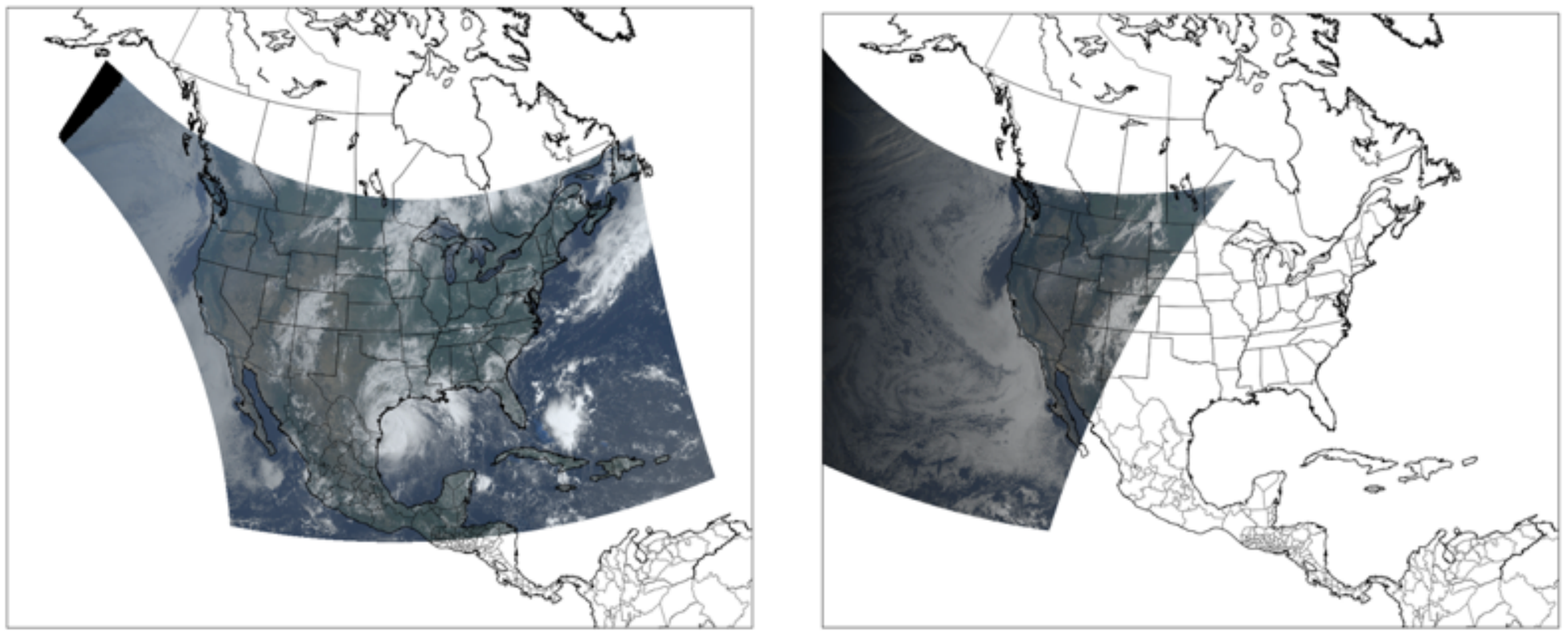}
 \captionof{figure}{GOES-16 (left) and GOES-17 (right) coverage area. We use GOES-16 satellite data as it observes the continental U.S.~ covering all the 25 solar sites considered in this work.}
\label{fig:g16g17wholedisk}
\end{minipage}%
\quad
\begin{minipage}[t]{0.48\linewidth}
\centering
\includegraphics[width=\textwidth]{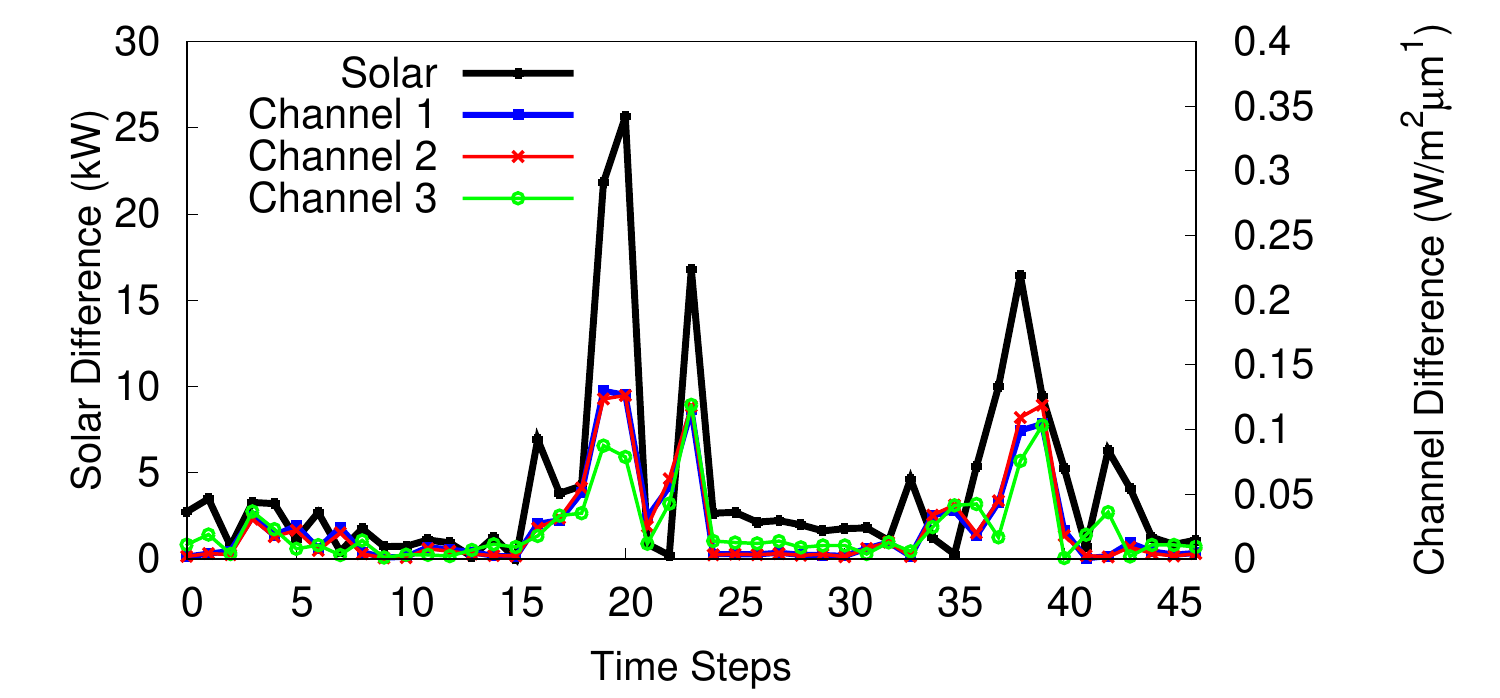}
 \captionof{figure}{Absolute differences between consecutive instants sampled at every 15 minutes. We show the solar PV output from a home and the first three spectral channels at that location. The differences in solar output are well tracked by differences in satellite channels.}
\label{fig:15minsdifference}
\end{minipage} 
\end{figure}



The GOES-R series of geostationary satellites started launching in late 2017, and now provides remote sensing data from 16 spectral channels that comprise different ranges of wavelengths of light, as well as numerous secondary derived products, such as estimates of downward shortwave radiation (DSR).  Note that solar PV only generates power from the first 3 of these spectral channels, which are mostly in the visible range of light. As a result, our work only trains models on these first three channels.  GOES-16 covers the entire U.S., while GOES-17 provides additional coverage for the Western U.S. and the Pacific ocean, as shown in Figure~\ref{fig:g16g17wholedisk}. Thus, we will use data from GOES-16 as it covers all the solar sites used in this work. The data has both high temporal and spatial resolution, including new spectral readings every 5 minutes for every 0.5-2km$^2$ area in the U.S., and is made publicly available in near real time.  The data is a rich source of information about the environment and climate that is useful for a wide range of applications. Solar forecasting is a particularly compelling application, since solar output correlates directly with the amount of light (of certain wavelengths) that reaches the ground.  


\subsection{Prior Work}

Traditionally, solar forecasts depend on some measure of cloud cover to assess the effect of clouds on solar output.  Cloud cover is commonly measured by weather stations in units of ``oktas,'' where 1 okta means that one-eighth of the sky is partially covered by clouds.  Oktas are measured at ground-level using sky mirrors. Unfortunately, oktas are a coarse and imprecise measure of cloud cover that is typically released by weather stations every hour.   In addition, not every solar site is located near a ground-level weather station that reports oktas.  Thus, even though cloud cover measurements in oktas are widely available, this data remains an unreliable and inaccurate basis for solar forecasting. 

A better basis for solar forecasting is direct ground-level readings of solar irradiance. The U.S. operates the Surface Radiation Budget Network (SURFRAD) ~\citep{SURFRADANationalSurfaceRadiationBudgetNetworkforAtmosphericResearch} within the U.S., which measures and records ground-level solar irradiance at different monitoring sites. These monitoring sites operate in collaboration with the National Oceanic and Atmospheric Administration (NOAA) that also release values of Downward Shortwave Radiation (DSR). Unfortunately, while SURFRAD measurements are precise, they are not widely available, as there are only eight SURFRAD sites maintained in the entire U.S. As a result, we cannot use SURFRAD data as a basis for solar forecasting models. Finally, NOAA releases derived data products from raw GOES spectral data, including DSR ~\citep{dsr}, which is an estimate of ground-level solar irradiance.  Unfortunately, satellite-based estimates of DSR are only released every hour and are often not released during cloudy conditions (when they are most important)~\citep{dsr-document}.  As a result, satellite-based estimates of DSR are also not a reliable basis for solar forecasting.  Instead, our work leverages a ML regression model that infers solar output directly from the spectral data, specifically using the channels in the visible range, as the basis for solar forecasting. 


 


Recent work \citep{dong,dsr2020,pvlib,buildsys,bashir2019solar} has focused significantly on solar performance modeling---inferring current solar output from current environmental conditions---but not forecasting. Solar forecasting is a much more challenging modeling problem, since it must infer, not only how spectral data correlates with a site's solar output, but also how the spectral data will change over time based on the movements of clouds.  Accurately forecasting hours, or days, in the future is challenging because of non-linear atmospheric dynamics that affect cloud movements, and which are not directly captured by GOES data. Such long-term forecasting on the order of multiple hours or days instead require Numerical Weather Prediction (NWP) algorithms \citep{richardson1922weather} that leverage non-linear physical models of the atmosphere, and require more inputs beyond spectral satellite data.  

On the other hand, near-term solar forecasting from satellite data is more tractable, since over short time periods, cloud movements are more heavily impacted by recent phenomenon.  As a result, models that incorporate historical spectral data across a region have the potential to track changes in the data over time as they move across a region.  Prior work on cloud motion vectors \citep{lorenz2004short,lorenz2012prediction} has taken this approach in identifying clouds and tracking cloud movements to assist solar nowcasting. However, they largely use physical models, and do not leverage either the latest multispectral data from GOES-R or recent advancements in forecasting using deep learning. The higher resolution data offered by GOES-R admits more accurate, localized, and near-term forecasts compared to prior work based on coarser and less precise data.  Similarly, recent advancements in deep learning offer a more automated ``black box'' approach that does not require manually calibrating physical models for specific data sources or solar sites. 

Finally, another line of research \citep{zhang2018deep,zhao20193d,siddiqui2019deep,paletta2020convolutional,nie2020pv} utilizes sky-images collected from specialized cameras installed directly at the solar site that continuously capture images of the sky at frequent time intervals. Since this requires specialized hardware to be installed at each solar site of interest, this severely limits the applicability of such models. Indeed, prior work on such methods has only considered a very small number of solar sites in their study, for instance, a maximum of only 2 solar sites in \cite{zhang2018deep,zhao20193d,siddiqui2019deep,paletta2020convolutional,nie2020pv}. In contrast, our work utilizes satellite data that is readily available for any location and thus our approach is applicable to any site in the continental U.S.A. We will demonstrate the utility of our solar nowcasting models on 25 solar sites, an order of magnitude larger than that considered in any of these prior methods.

\subsection{Basic Methodology}
As we discuss in \Section\ref{sec:design}, our approach breaks the solar nowcasting problem into two steps: (1) learning a self-supervised model on the raw satellite data across all sites of interest, and (2) modeling the future solar output at a particular solar site by utilizing predictions from the self-supervised model.
The self-supervised model combines a convolutional neural network (CNN) with a long short-term memory (LSTM) for time-series forecasting of spatial multispectral satellite data. CNNs are typically used for analyzing spatial imagery. Multi-spectral satellite data across a region of some size at any moment in time is akin to an image, where the spectral data is equivalent to a pixel value. In contrast, LSTMs have feedback connections that make them well-suited for forecasting temporal data, but cannot be directly applied to spatial data. Thus, combined CNN-LSTMs are useful when the input data has both a spatial and temporal structure, as in solar nowcasting. 

Specifically, 
given a sequence of spectral data covering some area over some number of previous time steps, the self-supervised model is trained to predict the value of the spectral data at the center location corresponding to the solar site's location in the next time step.  In our case, we focus on time-steps of 15 minutes in particular.
Note that this a self-supervised model as it only makes use of the raw spectral data to predict subsequent samples in that datastream that is already captured by the GOES-16 satellite and does not require any site-specific PV generation data from installed solar sites. 
This is akin to self-supervised models in ML literature, for instance, language models \citep{radford2019language} that predict the next word in a sentence or general auto-regressive models that predict future samples in a sequence \citep{oord2018representation}. These models can be further specialized to specific supervised tasks of interest.

In order to enable site-specific solar nowcasting, the predictions of the future spectral data at the site's location from the self-supervised model can then be fed to a regression model that infers a site's solar output from given spectral data. These models are capable of incorporating site specific information that affects solar output from current conditions, such as a site's size, orientation, tilt, and shading from obstructions, as we discuss in \Section\ref{sec:design}.

\begin{figure}[t]
\begin{center}
\centerline{\includegraphics[width=0.6\textwidth]{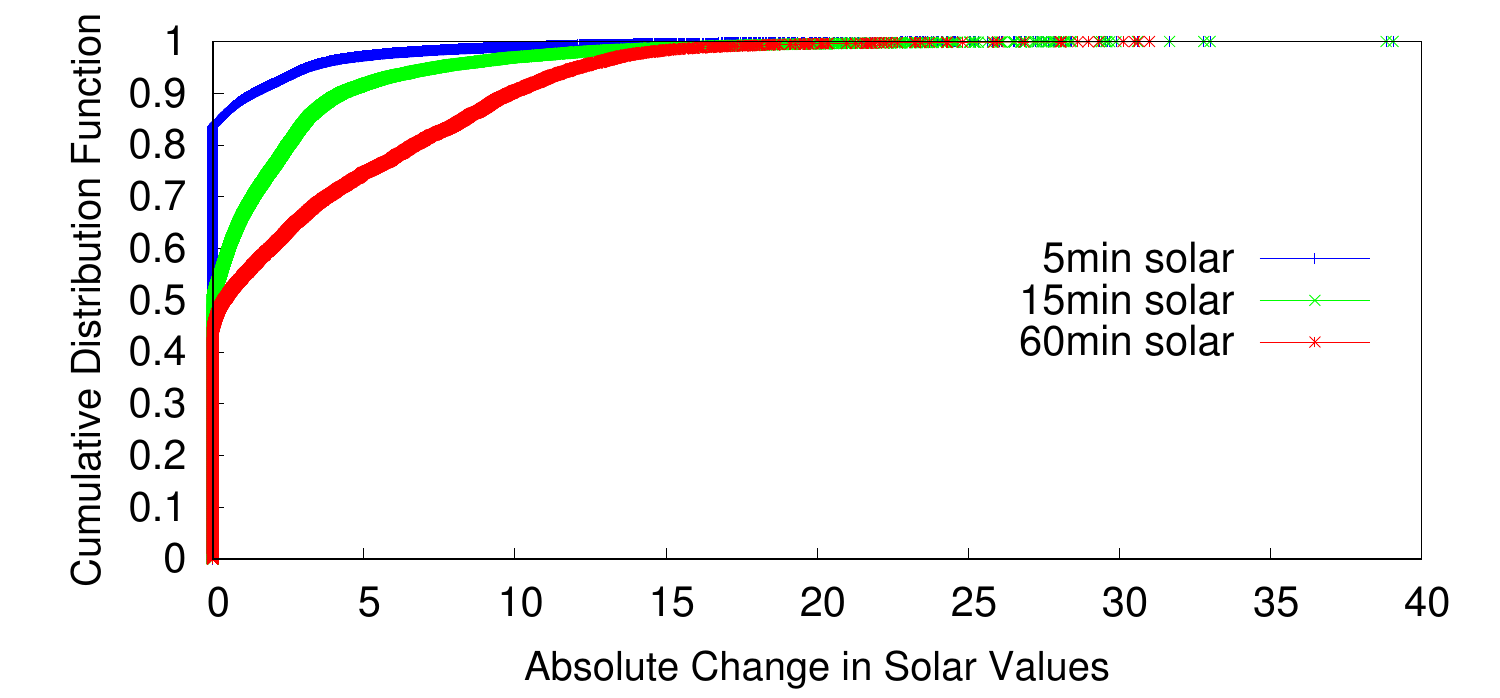}}
\caption{Cumulative Distribution Function of changes in solar output at 5, 15 and 60 minute frequency over one year for a representative solar site. Note that a majority of times solar hardly changes between instants (value at $0$) and thus modeling the occasional abrupt changes is crucial.}
\label{solarcdf}
\end{center}
\end{figure}

We evaluate our solar nowcasting models above in \Section\ref{sec:evaluation} across 25 solar sites over a year.  Our evaluation particularly focuses on the accuracy of our models to predict large changes in solar output over short time periods.  These are the changes that are potentially most disruptive to solar-powered systems, including the electric grid.  In addition, evaluating solar nowcasting over all time periods obscures the problem, since solar output often does not change much within a 5-15 minute period.  As a result, a simple persistence model that predicts solar output never changes over 5-15 minutes are highly accurate when averaged over many time periods, even though they are highly inaccurate, by definition, when any change in solar output occurs. Figure~\ref{fig:15minsdifference}  illustrates this point by showing the change in solar output every 5 minutes, as well as the first three spectral channel values, over a day for a particular solar site.  As shown, most of the time, neither the solar output nor the channel value changes significantly.  However, there are a few times within the day that experience significant changes.  These significant changes are the ones that are most important to accurately predict, as they have the most potential to disrupt solar-powered systems and the grid.  This graph also demonstrates the correlation between the spectral channel values sensed by the satellite and a site's solar output: they tend to rise and fall in tandem, although the magnitude of the increase and decrease varies over time.  

Figure~\ref{solarcdf} then shows a Cumulative Distribution Function (CDF) of the change in solar at 5, 15, and 60 minute periods.  This graph shows that, as expected, there are fewer large changes in solar output at small intervals, and the size of the changes are generally larger over longer periods.  Thus, accurately predicting the few large changes can be a challenging problem.  As a result, we condition our evaluation in \Section\ref{sec:evaluation} on the accuracy of predicting changes in near-term solar output above a specified magnitude. 
Moreover, note that at 5-minute interval, close to 80\% of the data (value at $0$) have no change in solar in subsequent time steps. Thus, we choose a 15-minute interval for our study, so that there will be more instances with non-trivial changes in solar.

%% file: design.tex
\begin{figure*}[t]
\begin{center}
\includegraphics[width=\linewidth]{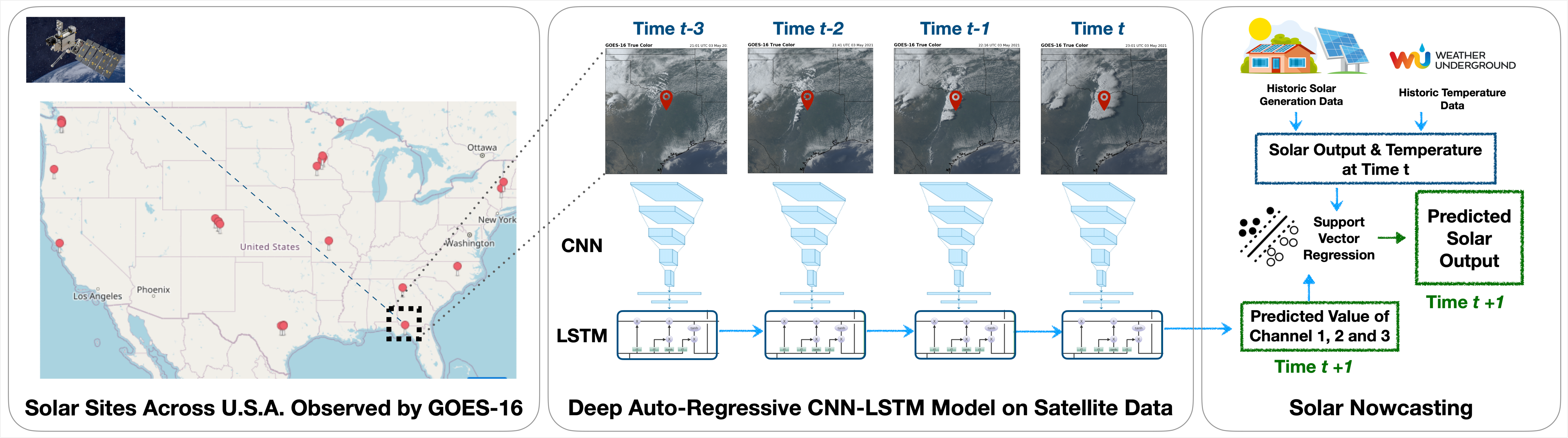}
\caption{Overview of our complete pipeline. Multispectral satellite data from 25 sites across U.S.~ is collected. Sequences of observations at a site, comprising of first 3 channels from the satellite (visualized in middle) are input to a CNN-LSTM model that is trained to predict the observation at the center (red pin) for the next time step ($t+1$). An auto-regressive SVR takes the predicted channels and previous solar output as well as previous temperature to predict the solar output at next step.}
\label{design}
\end{center}
\end{figure*}

In this section, we describe our methodology in developing a solar nowcasting model using multispectral satellite data.   Figure~\ref{design} shows an overview of our modeling approach. We first describe our neural network model for self-supervised learning on raw satellite observations, and then describe a simple auto-regressive model for forecasting satellite channel values that can leverage forecasts from the self-supervised model.   We then feed these forecasted channel values into a regression model that infers a site's solar output. 

Given a set of solar sites of interest, we consider an area of $w \times w$ around each site, and extract the 3 visible channels of satellite data from the GOES-16 satellite. Note that each location $l$ within the area is described by a 3-tuple of channel values.  We denote each such 3D image of dimensions $(w \times w \times 3)$ at a location $l$ observed at instant $t$ as $I_t^{(l)}$.  We then extract a temporal sequence of these images over time from the satellite, with the target site $l$ always at the center of the image as this is a geo-stationary satellite.  This data effectively has four dimensions described by the 2D area (length and width), three channel values, and time.  We vary both the area and the amount (and resolution) of historical data we use for training in our models, as the optimal values depend on a location's climate and the target forecast horizon. For example, we could use the previous 4 satellite images for a region of size $w$ with an interval of 15 minutes between images, or the last hour of changes in the spectral data.   Larger areas and longer historical time periods increase the training data size, which increases the computational overhead of training.  As we discuss, however, there are diminishing returns with respect to improvements in accuracy as these values increase.

\subsection{Spatial Modeling using Convolution Neural Network (CNN)}

Data from the first three channels over an area from the satellite forms a 3D image $I_t^{(l)}$ that we first process using a Convolution Neural Network (CNN) to extract spatial features from the image.  A CNN model \citep{lecun1989} is the standard neural network architecture used for modeling visual imagery and extracting visual features. CNN models comprise of trainable convolution filters and pooling operations that together extract spatially invariant features from images. We use multiple layers of convolution filters followed by max pooling layers. The exact CNN architecture is described in \Section\ref{sec:hyper}. 
The output of processing the image with the CNN model is a $k$-dimensional feature vector:

\vspace{-0.2cm}
\begin{equation*}
    v_t^{(l)} = CNN(I_t^{(l)}; \theta)
\end{equation*}

where $\theta$ represents trainable parameters of the CNN model and $v_t^{(l)}$ is the extracted $d$-dimensional spatial feature vector for location $l$ at time $t$. Note that the CNN is trained to extract spatial features that help model the temporal dynamics of the satellite data as explained subsequently.

\subsection{Temporal Modeling using Long Short-Term Memory (LSTM)}

Short-term forecasting of solar power needs to take into account the recent history of changes to solar irradiation at the surface and how it will evolve in a short span of time.  We use long short-term memory networks (LSTM) \citep{hochreiter1997long} to capture the evolution of the per-instant spatial features extracted from the CNN over time, which is crucial for predicting future satellite channel values. The LSTM is a prominent neural architecture used for modeling sequences of data and is also often employed in time-series forecasting. LSTM makes use of a cell state, that is an internal memory summarizing the previous history at a given time, and a hidden state that's the output of the current time step. Multiple gating mechanisms update the cell state by combining it with the current input and the previous hidden state. In our case, the LSTM update at step $t$ is summarized as:
\begin{equation*}
    s_t^{(l)}, h_t^{(l)} = LSTM(s_{t-1}^{(l)}, h_{t-1}^{(l)}, v_t^{(l)}; \phi)
\end{equation*}
where $\phi$ are the LSTM trainable parameters, $s_t$ is the LSTM cell state, $h_t$ is the hidden state, and $v_t^{(l)}$ is the CNN spatial feature vector at time $t$ for location $l$. Thus recursively reapplying the same function at every time-step, the LSTM models the evolution of the input features over time.

After processing a sequence of $T$ images through the CNN and LSTM, for instance the satellite imagery over the previous one hour, the final hidden state $h_T$ of the LSTM summarizes the entire sequence.  This state is passed through a dense layer with sigmoid output units to predict the value of the visible channels at the site's location for the next step:
\begin{equation*}
    \hat{C}_{T+1}^{(l)} = \sigma\left( W h_T^{(l)} \right)
\end{equation*}
where $W$ is a $(3\times k)$ matrix with $k$ being the hidden state dimension and $\hat{C}_{T+1}^{(l)}$ are the predicted values of the 3 channels of satellite at the next time instant.

\subsection{Training CNN-LSTM Model}

The CNN-LSTM model is trained end-to-end in an auto-regressive manner. That is, given the sequence of past images we use the model to compute the predicted values for the next instant and use mean-squared error  with respect to the true future satellite values as the loss function:

\begin{equation*}
    \min_{\theta, \phi} \sum_{l, T} \left\lVert \hat{C}_{T}^{(l)} - C_{T}^{(l)} \right\rVert^2
\end{equation*}

where $C_{T}^{l}$ is the ground-truth satellite observations for all 3 channels at time $T$ and location $l$, $\hat{C}_{T}^{l}$ is the prediction from the CNN-LSTM model as described above, and $\lVert\cdot\rVert$ represents the euclidean norm.
Note that the prediction is a function of both the CNN spatial extraction model and LSTM temporal extraction model such that backpropagation optimizes the parameters of these models to extract features that can predict future observations well.
Thus, using widely available satellite data, we train CNN spatial extraction and LSTM temporal models to capture the dynamics of the multi-spectral satellite data.  Note that, in our evaluation, we only train one global model by combining satellite data across multiple sites.  This enables modeling of shared statistical properties across sites rather than overfitting to the peculiar characteristics of an isolated site.  Moreover, this also provides a large amount of data for learning a useful CNN-LSTM model which typically do not work well with small datasets.

The drawback of global modeling is that it does not account for unique aspects of any specific location's climate.  A local model trained only on data from a specific location is capable of identifying such unique attributes to improve accuracy.  For example, in some locations, winds may typically move west to east, while in others, they may typically go in the other direction.  However, training our models requires a significant amount of data, and, since the GOES-R satellites only began releasing data a few years ago, there is not yet a large volume of data available for training and testing on any single location.  In addition, a global model is beneficial because it does not require re-training, and can be applied to any location.

%% file: model_solar.tex
\label{sec:solar}
As we discussed earlier that the GOES visible bands are highly correlated with solar irradiance at the surface, this allows accurate inference of solar power output through machine learning models trained on historic generation data with GOES visible bands as inputs. We seek to utilize the same relationship (see Figure~\ref{fig:15minsdifference}) for solar forecasting by using deep neural networks to implicitly model the short-term changes in the values of the visible bands of GOES. 
We use the CNN-LSTM model from the previous section to forecast future values of the visible channels which in turn will help in predicting solar generation, owing to this relationship between channel value and solar generation.

Given a trained CNN-LSTM model that can generate future satellite observations at a given site, we leverage these predictions in a model for solar power forecasting at any solar installation site of interest.
We leverage this relationship between visible bands and solar irradiance by considering the following auto-regressive model for forecasting near-term solar output:
\begin{equation}
    P^{(l)}_{t+1} = f(P^{(l)}_t, C^{(l)}_{t+1}, T^{(l)}_{t}) \label{eq:solar_model}
\end{equation}
where $P_t$ is the solar power generated, $C^{(l)}_{t}$ are the satellite channel values and $T^{(l)}_{t}$ is the temperature at time $t$. $f(\cdot)$ is a regression model, such as support vector regression, that models the relationship between the input and output variables using historical data.
Temperature is an important component of solar generation as solar panel efficiency is sensitive to the surrounding temperature \cite{dong}.
Note that we use $C^{(l)}_{t+1}$ instead of $C^{(l)}_{t}$ in \eqref{eq:solar_model}. 
A major component of change in $P_{t+1}$ from $P_{t}$ is captured in the change in $C_{t+1}$ from $C_{t}$. This complex relationship is modeled using our CNN-LSTM model, described above, which allows us to predict an estimate $\hat{C}^{(l)}_{t+1}$, which is an estimate of true channel values at $t+1$ for the auto-regressive model in \eqref{eq:solar_model}.

The regression model for solar nowcasting in \eqref{eq:solar_model} is trained using current satellite observations. That is, $f(\cdot)$ is trained using the ground-truth satellite observations at $t+1$ time-step ($C^{(l)}_{t+1}$), historic solar output at the location $l$, and the historic temperature data at location $l$. This step does not require the use of the CNN-LSTM model.
This is required to train an accurate auto-regressive model that, given true satellite observations, can learn to correctly infer future solar output. 
Once the regression model is trained, instead of using true future satellite observations, which are unavailable, we replace it with estimates from the CNN-LSTM model ($\hat{C}^{(l)}_{t+1}$) to provide forecasts.

%% file: implementation.tex

\subsection{Satellite data and solar sites}
The GOES-16 multispectral data is made publicly and freely available by NOAA as netCDF files hosted on Amazon S3 buckets. We use a script to recursively download the data for each date each year along with the description of the ABI product, bucket, domain, and the satellite name. The size of each 5 minute netCDF file is in the range of $\sim$75MB, which requires nearly 16 terabytes to store two years of data from a single GOES-R satellite.  Each 5 minute file includes data across 16 spectral bands covering the entire American subcontinent. To minimize storage requirements, we filter each file as we download it to extract only the relevant channel data for specific area around the locations of interest, and discard the rest.  

The netCDF files for multispectral data require some processing to filter out the data for the location of interest.  Specifically, our python modules read the \textit{goes\_imager\_projection} variable to convert $(x,y)$ degree coordinates for latitude and longitude to radians. We then search the file for the latitude-longitude pair that is closest to our location of interest. Since these are geostationary satellites, their rotation matches that of the Earth, enabling us to look at the same part of the file each time.  Thus, we read a single netCDF file and first create a list of the closest latitude-longitude pairs using the Vincenty formula~\citep{vincenty1975direct}, which calculates the distance between two points on the surface of a spheroid. This is done to reduce computational resources so that the process of finding the nearest location is not repeated for each 5 minute file. 

For all of our analysis, we use data for 25 solar sites across two years. 
Satellite data for the continental U.S. is extracted for the 2019 year. We restrict modeling to a $10\times10$ window around the 25 solar sites which constitutes the training data for the CNN-LSTM model for compute efficiency. The $10\times10$ window covers an area of approximately 10km$^2$. We average observations in a 15-minute window, which  reduces the sequence length for modeling and noise in the data by reducing the number of missing observations and sensor errors.
Solar sites used in this work can be seen in Figure~\ref{design} and are uniformly spread across the continental U.S., including the two coasts and central regions.
Since solar modeling only makes sense during the day time, we restrict the satellite data to be from 9am in the morning to 5pm in the evening based on the local time of the solar sites.
This yields more than \num[group-separator={,}]{300000} 5-step sequences of $10\times10$ images (with 3 channels) at intervals of 15 minutes.

We use 5-fold validation in all the experiments, splitting by day so that test sets have entire days held out for evaluation. 
This is done for both types of evaluation: evaluating channel prediction models and evaluating end-to-end solar nowcasting.
Table~\ref{table:trainingpointsetc} shows the training, validation, test split of these satellite observations used in this work.
Solar generation data from the energy meters for the same 25 sites and temperature data from the weather station are obtained from years 2018-19. We restrict generation data to be from 9am to 3pm every day, which is the peak duration of solar generation.

\subsection{Model hyper-parameters and metrics}
\label{sec:hyper}
CNN model comprises of 2 blocks of convolutions, where each block contains 2 convolution layers with 32 filters of size $3\times3$ and ReLU activation followed by a max-pooling layer of size $2\times2$. This is followed by two dense layers with hidden dimension $k=256$ and ReLU non-linearity between layers.
We use a one layer LSTM that takes these 256 dimensional inputs and has a hidden state dimension of 64.
Hyper-parameters for this and other ML models considered (decision tree and random forest) were determined using the validation set.

We use two metrics to evaluate the performance of our channel forecasting models and end-to-end solar forecasting models. Mean Absolute Percentage Error (MAPE), which quantifies the average percentage across time. 
\begin{equation*}
    MAPE =  \frac{1}{n} \sum_{t=0}^{n} | \frac{A_{t} - P_{t}} {A_{t}} | \quad\quad
    MAE = \frac{1}{n}\sum_{t=0}^{n} |A_{t} - P_{t}|  
\end{equation*}

Here, $A_{t}$ and $P_{t}$ represents the actual and predicted values. 
MAPE, which is often used to quantify the performance in prior work \citep{wang2019review}, is an intuitive metric and is comparable across solar sites of different installation sizes and configurations. 
However, it is sensitive to periods of low absolute solar generation and can be significantly affected by small absolute errors. We also used mean absolute error (MAE) to quantify the error in channel modeling given that the first three channels are reflectance values in the range of 0 to 1.
In the graphs, we use AE times 100 to plot unless otherwise stated. 

%% file: evaluation.tex
\begin{table}[t]
\small
\begin{center}
\resizebox{0.6\columnwidth}{!}{
\begin{tabular}{|l|c||c|}
\hline
\textbf{\emph{Data Sets}} & \textbf{\emph{ Number of points}} & \textbf{\emph{ Number of days in a year}} \\
\hline
\hline
Training & 236375 & 262 \\ \hline 
Validation & 27040 & 30 \\ \hline  
Testing & 65258 & 73 \\ \hline  
\end{tabular}
}
\end{center}
\caption{Total number of sequences as well as the number of days of the year that comprise training, validation and testing for the CNN-LSTM model.}
\label{table:trainingpointsetc}
\end{table}

In this section, we evaluate our proposed CNN-LSTM for deep auto-regressive modeling on satellite data and its utility for end-to-end solar nowcasting.
First, we evaluate the efficacy of the CNN-LSTM model for predicting future values of satellite channels for a given location. We consider our evaluation along spatial and temporal axes, as well as consider alternative ML models.
Then, we utilize the trained CNN-LSTM model obtained from that analysis for solar nowcasting at the site locations.
We use the following terminology throughout the evaluation:\\

\noindent \textbf{Persistence Model}: Since all the models considered here predict values for the next instant, typically 15 minutes in the future, a natural baseline is one that assumes there will be no change in the predicted quantity also termed as the persistence model. 
As we discussed earlier, solar output often changes in small, abrupt bursts and thus a large fraction of the time there is negligible change in near term solar output (see Figure~\ref{solarcdf}). Thus, improving over this persistence model prediction is difficult and is of interest in this work.\\

\noindent \textbf{Tolerance}: As across the year, changes in solar (and hence satellite observations) over short durations like 15 minutes are often negligible, we perform all of the analysis in buckets of varying threshold over subsequent changes.
That is, we take a tolerance $\delta$ and consider only points $x_i$ where subsequent changes were at least $\delta$: $\{x_i |\; |x_{i-1} - x_i| \ge \delta\}$.
We evaluate all models over a range of different values of $\delta$ to provide a sense of how they perform over both small and big, sudden changes in solar.
Table \ref{table:tolerancepoints} lists the fraction of points in the validation data for each tolerance value considered. \\

\noindent \textbf{Forecast Skill Score}:
We also use "forecast skill score" (SS) to compare the performance between various methods, which is also used in prior work \citep{zhang2018deep,wang2019review}, given by:

\begin{equation*}
    SS = \left(1 - \frac{\mathcal{E}_{\mbox{prediction}}} {\mathcal{E}_{\mbox{baseline}}}\right) * 100\% 
\end{equation*}

Here, $\mathcal{E}$ is the error metric used to evaluate performance for every model. If the \textit{prediction model} performs equally well as the \textit{baseline model}, the skill score will be 0. A higher skill score thus means that the prediction model outperforms the baseline model. We will use the skill score to compare the performance between different models. For the skill score, the baseline is always the persistence model as presented above.

\begin{table}[h]
\begin{center}
\resizebox{0.6\columnwidth}{!}{
\begin{tabular}{|l|c||c||c|}
\hline
\textbf{\emph{Tolerance}} & \textbf{\emph{ \% of points C01}} & \textbf{\emph{ \% of points C02}} & \textbf{\emph{ \% of points C03}} \\
\hline
\hline
0 & 100 & 100 & 100 \\ \hline 
0.01 & 65.24 & 68.33 & 80.56 \\ \hline  
0.02 & 47.32 & 49.95 & 64.54 \\ \hline  
0.05 & 22.0 & 24.34 & 35.39 \\ \hline 
0.10 & 8.38 & 9.63 & 13.29 \\ \hline 
\end{tabular}
}
\end{center}
\caption{Variation in number of data points with respect to different tolerances for different channels.}
\label{table:tolerancepoints}
\end{table}

\subsection{Evaluating ML models for spatial modeling}

\begin{figure}[t]
\begin{minipage}[t]{0.48\textwidth}
\begin{center}
\includegraphics[width=\textwidth]{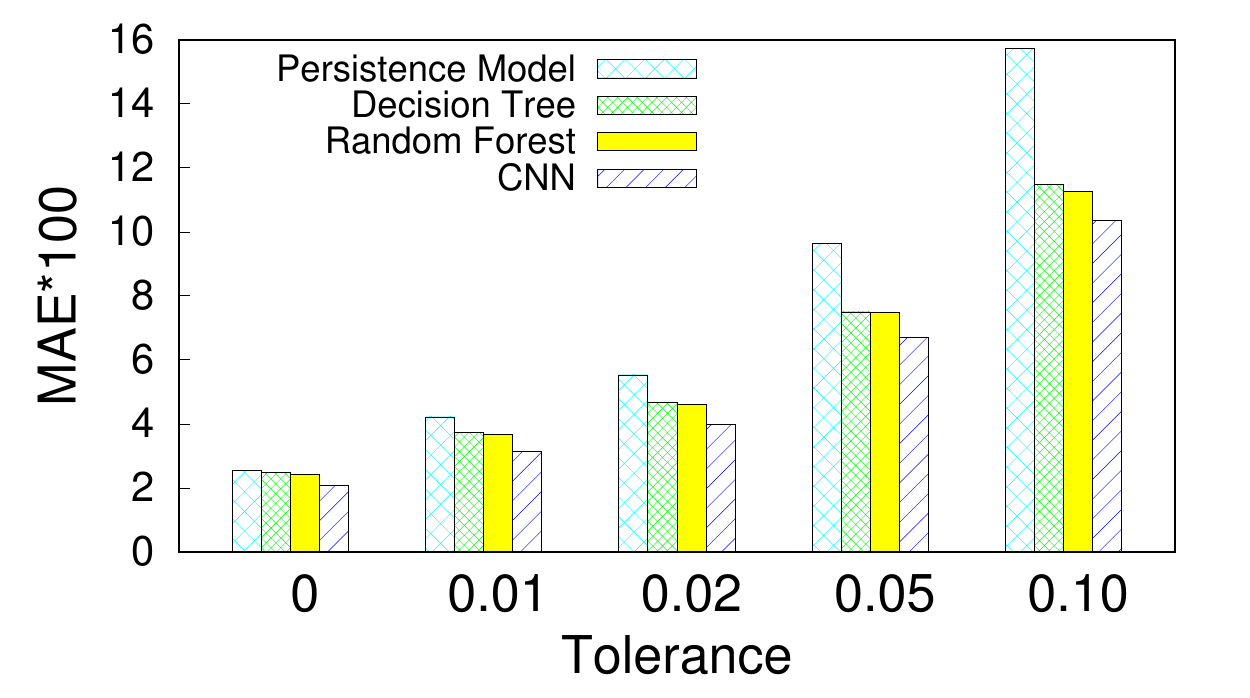}
\caption{MAE for different ML models used to predict the next instant channel values from current instant. CNN model has the lowest error.}
\label{mltolerance}
\end{center}
\end{minipage}
\quad
\begin{minipage}[t]{0.48\textwidth}
\begin{center}
\includegraphics[width=\textwidth]{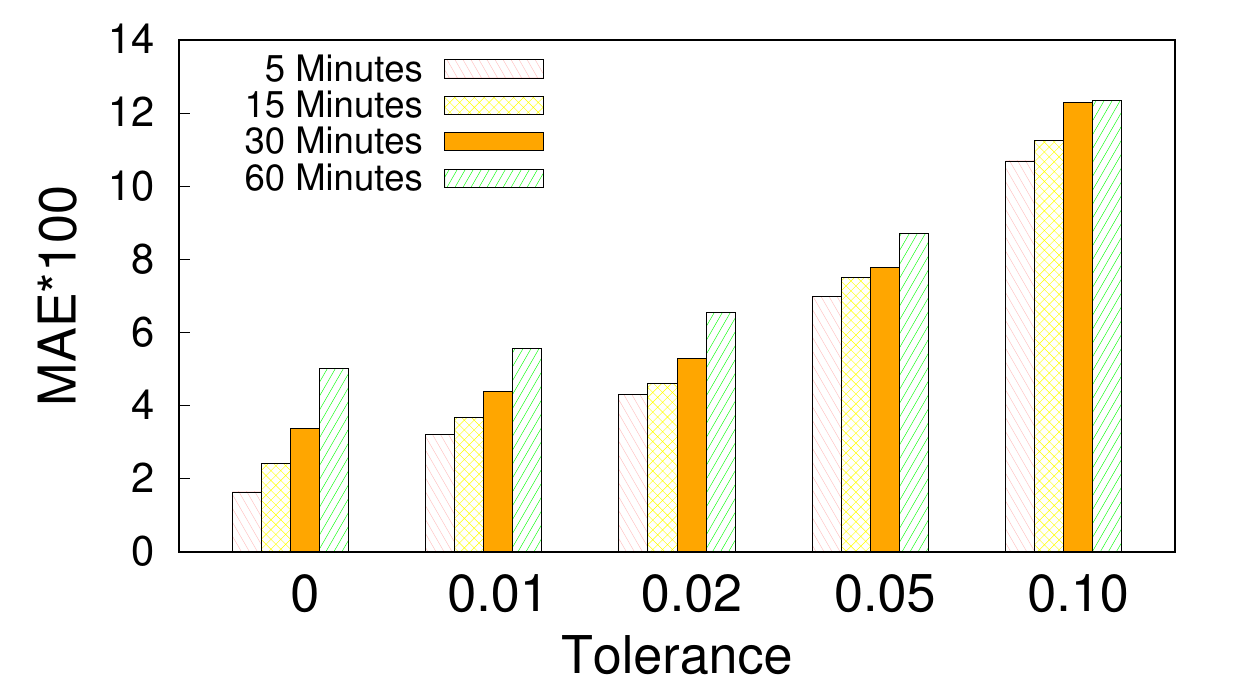}
\caption{MAE for models trained/evaluated at different time intervals. At larger intervals (30-60 minutes) forecasting becomes increasingly harder.}
\label{rftimetolerance}
\end{center}
\end{minipage}
\end{figure}


In this section, we consider different choices of ML models for spatial modeling and evaluate their utility compared to using CNN. We also consider choices like the spatial area and its effect on predicting future channel values.
For this purpose, we consider three standard ML models: (1) Decision Tree, (2) Random Forest, and (3) Convolution Neural Network (CNN).
Note that we only analyze next step predictions given current data (that is a lag-1 time-series models) and consider longer history temporal modeling in the next section.
We only show results for channel 1 to avoid repetition as results for other channels are qualitatively similar.
To train our decision tree and random forest models, we flatten the $w\times w$ spatial satellite observation into a vector of size $w*w$ that is input to the model.

Figure~\ref{mltolerance} shows the mean absolute error (MAE) for all three models at different tolerances. Here all models are trained over a $10 \times 10$ area around each solar site and predict the satellite observations in the next 15 minutes. The point of this graph is to show how the accuracy of a deep learning approach improves relative to that of simpler non-spatial models as the size of the subsequent change increases.  At $0$, which represents all of the data points, the CNN model is only marginally better than the other models. However, this occurs primarily because most of the time there are only small changes in solar over short time periods, as evident from the low error of the persistence model.  As we increase the size of the changes we examine, we see that the persistence model predictions, which assume the past is the same as the future, become increasingly worse, while the CNN model remains the best and improves over the other models by a large margin. 

\begin{figure}[t]
\begin{center}
\includegraphics[width=0.7\textwidth]{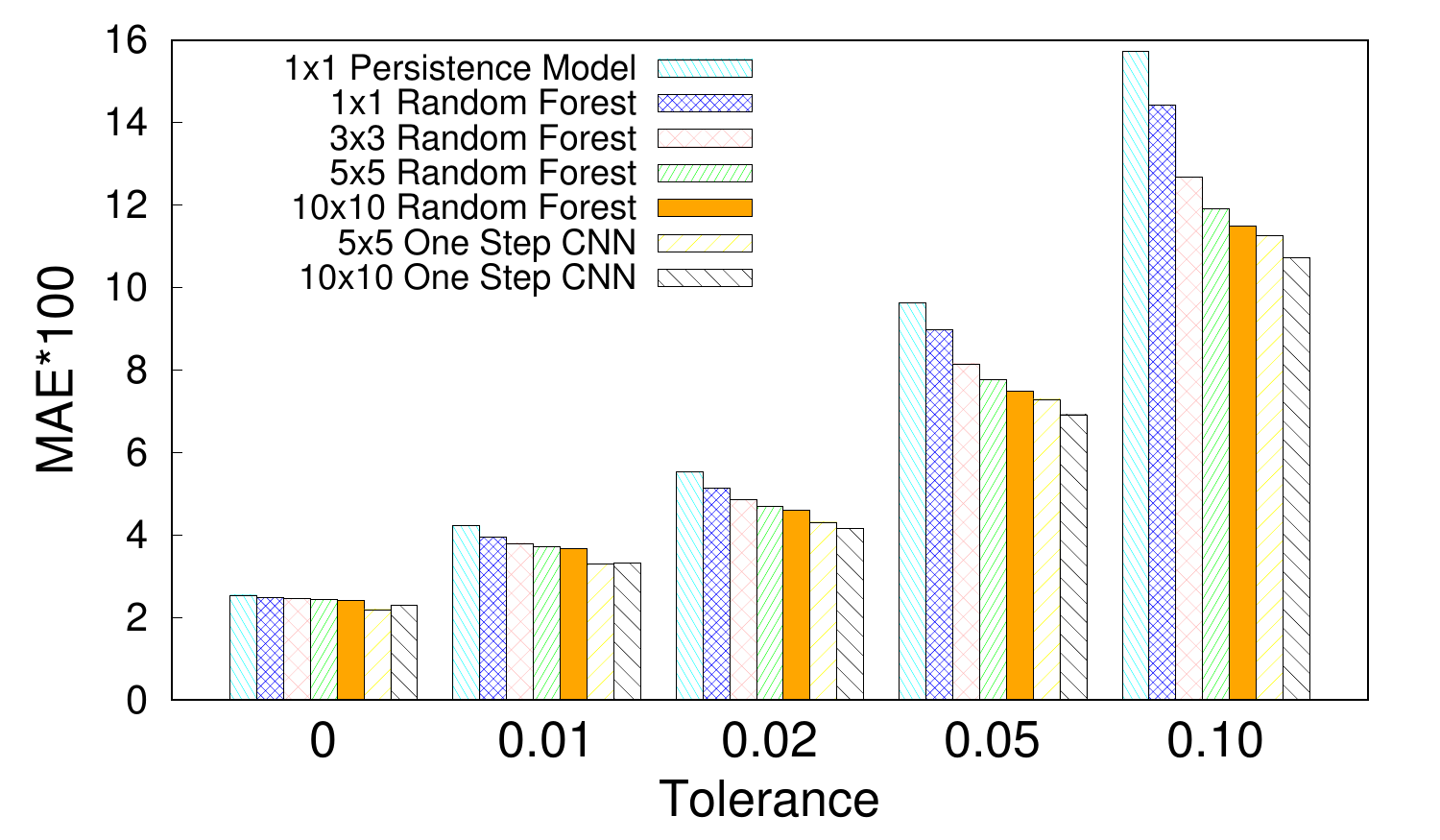}
\caption{Effect of spatial area on forecast MAE for different models. Using larger area improves forecast for all models and CNN trained using $(10\times10)$ area is better.}
\label{rftimearea}
\end{center}
\end{figure}

Figure~\ref{rftimetolerance} shows the effect on forecast error for models trained and evaluated at intervals of 5, 15, 30 and 60 minutes. We use the random forest model for this evaluation as training a CNN for every setting is expensive.
As expected, predicting further into the future is less accurate, since more changes occur.  This discrepancy in accuracy is most evident at $0$ tolerance when we include all data points.  This occurs because there are few changes in solar output over 5 minutes on average, while on average there are much more significant changes over 60 minutes, including changes due to movement of the sun in the sky. As we increase the tolerance to assess the accuracy of predicting larger changes, as expected, the error increases.  However, interestingly, the discrepancy in error actually decreases.  That is, the error in predicting a large change  30-60 minutes in the future is more similar to predicting a large change 5-15 minutes in the future. 

Figure~\ref{rftimearea} then compares the effect of using different sizes of spatial area, focusing on the two best models from Figure~\ref{mltolerance}. 
We train models with areas of $1\times1$ (i.e.~just the site), $3\times3$, $5\times 5$ and $10\times10$.
The graph shows that increasing the spatial area around the site used by the model results in a large improvement in accuracy. 
We see that using $10\times10$ area gives best results, and it is much better than not considering any surrounding area from the site ($1\times1$).
Moreover, using the CNN model results in much better spatial processing and improved results over variants of random forests.

\begin{figure}[b]
\begin{minipage}[t]{0.48\textwidth}
\begin{center}
\includegraphics[width=\textwidth]{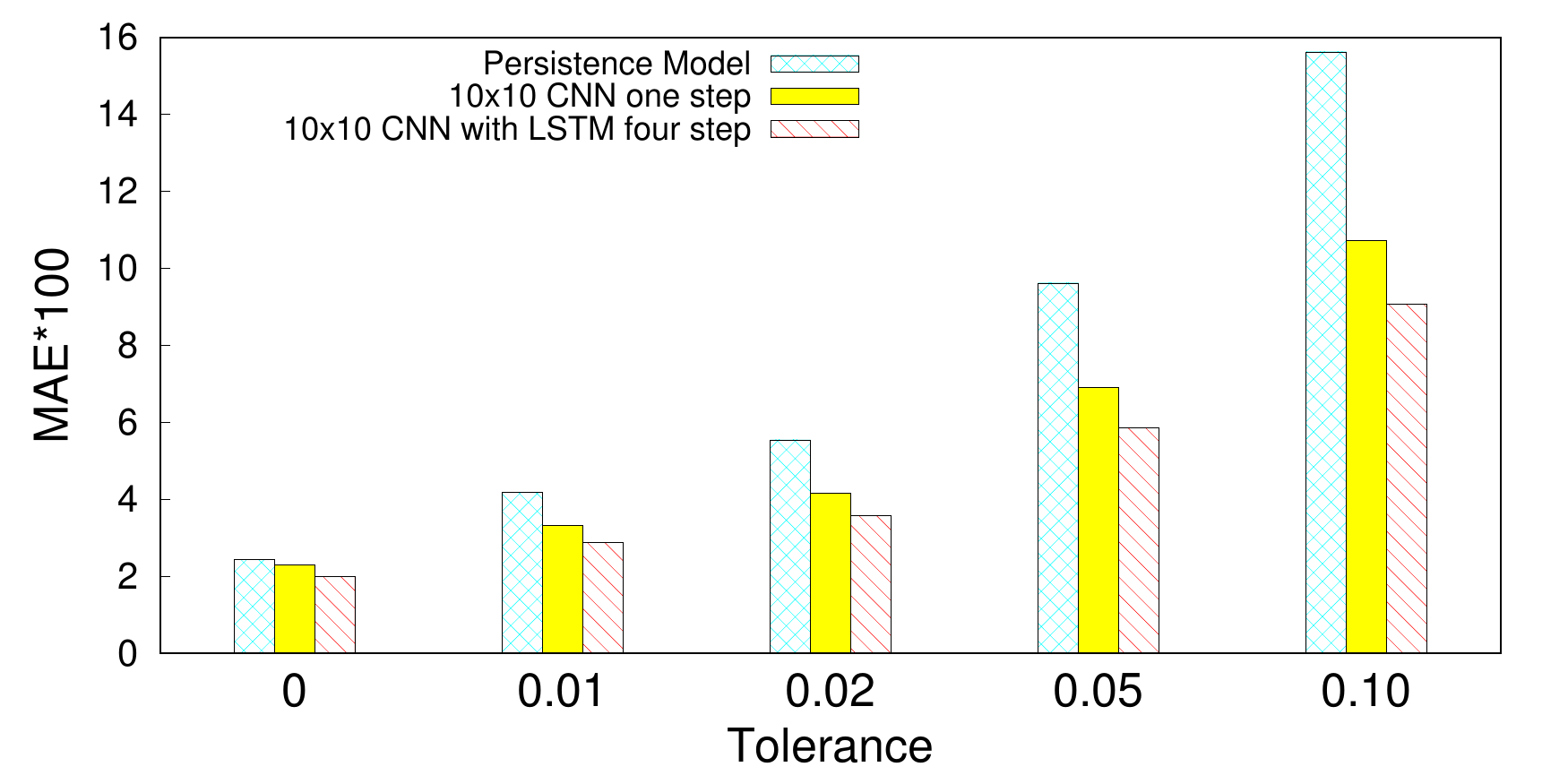}
\caption{CNN-LSTM performance at forecasting next instant channel values. Compared to not using previous temporal history (CNN one step), CNN-LSTM leads to significant error reduction on predicting large changes (tolerance$>0$) and retains overall better performance compared to persistence model (tolerance$=0$).}
\label{mlandcnn}
\end{center}
\end{minipage}
\quad
\begin{minipage}[t]{0.48\textwidth}
\includegraphics[width=\textwidth]{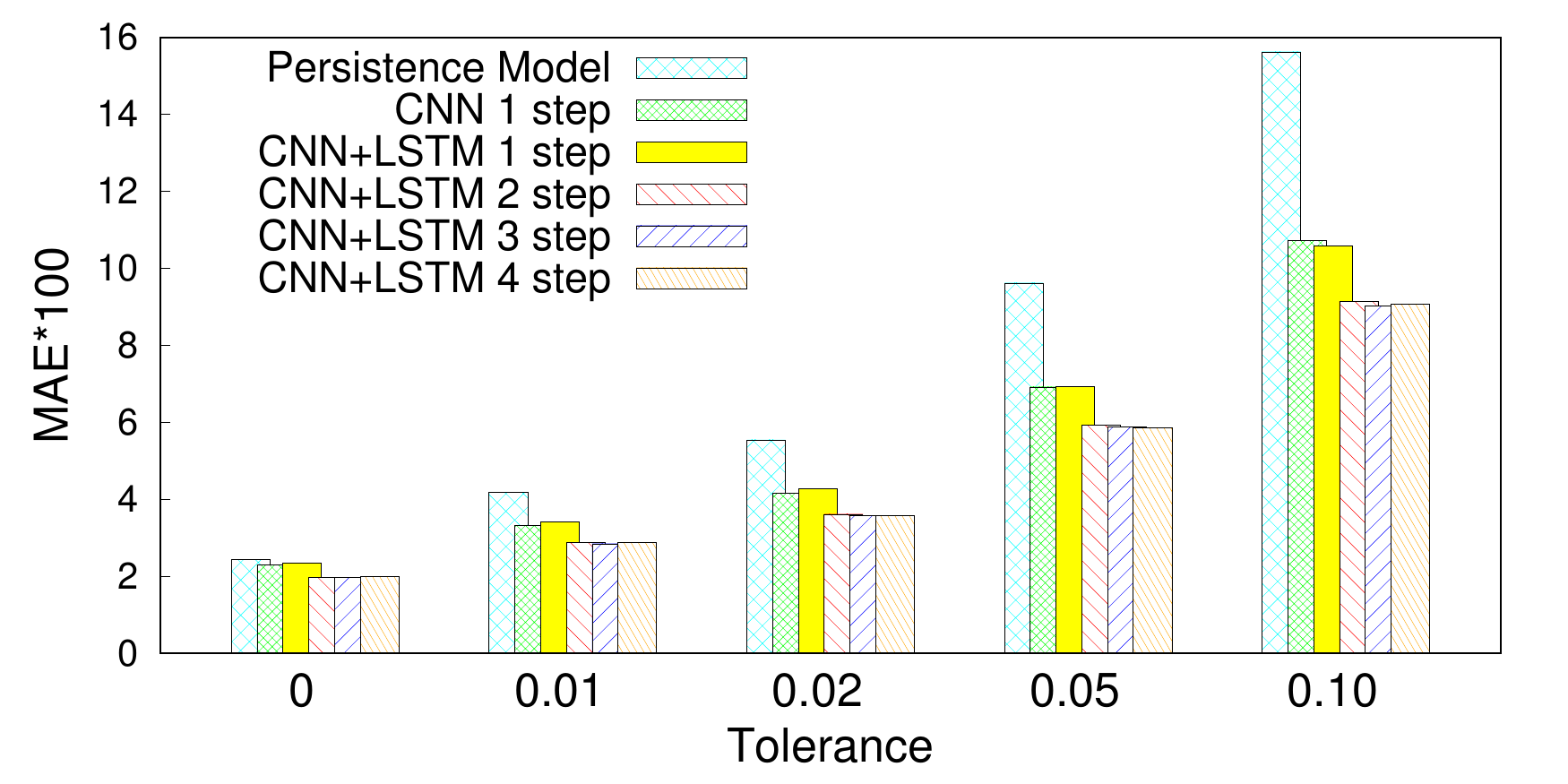}
\caption{Comparison of CNN-LSTM models using varying amounts of previous temporal information. Note that using 3-4 steps is better than using lesser steps. There is a marked reduction in error when using more than 1 step in the model. }
\label{fig:cnnlstmc0123}
\end{minipage}
\end{figure}


\subsection{Evaluating CNN-LSTM models for spatio-temporal modeling}

We now add the LSTM on top of the $10\times10$ area CNN model and utilize the previous timesteps as input to the CNN-LSTM model. In this case, we use 4 steps, which means we train the model on a dataset that includes the 4 previous $10\times10$ spatial regions, corresponding to the past 1 hour of spatial observations. 
The overall results are shown in Figure \ref{mlandcnn}, comparing the CNN-LSTM model with a single step CNN model using $10\times10$ spatial area.
The use of the LSTM drastically improves the results in terms of accurately predicting large changes.  Similar to the previous graph, when evaluating across all of the data (tolerance 0), the improvements over single step CNN are not significant because most of the time there are only small changes in solar output.  
However, the advantage of the CNN-LSTM becomes apparent when we look at predicting any significant change larger than 0 (tolerance 0.01) and as we increase the threshold of the changes we look at, we observe that our model that covers a $10\times10$ area and combines a CNN-LSTM leads to larger reduction in errors. 



We next evaluate the performance of CNN-LSTM variants in forecasting next time instant channel values.
We explore the following temporal variants: CNN using 1-step static image, CNN-LSTM using 1-step static image, CNN-LSTM using 2/3/4 steps of images in the past.
Figure~\ref{fig:cnnlstmc0123} shows our results compared with the precision model predictions.
Incorporating multiple steps of information in CNN-LSTM is better than using the current static image for forecasting, showing the utility of a deep auto-regressive approach.
We find that using 3 or 4 steps, i.e., 45 minutes or 60 minutes in the past, perform comparably. 
Notably, using even just 2 steps leads to a marked reduction in error. This signifies that the model is likely able to infer temporal changes in the satellite data, such as cloud movement, for better prediction.

\subsection{End-to-End Solar Nowcasting}

The end goal of this work is to use these channel predictions and translate them into an end-to-end solar nowcasting model. In this section, we evaluate the utility of our models for this purpose. Since we want a clear comparision of the benefit of using the self-supervised CNN-LSTM model for solar nowcasting, we use the CNN-LSTM model as a fixed model in the nowcasting part. That is, after the self-supervised learning on raw satellite observations, this model is fixed and not trained further on any site-specific data from the solar installation sites. This enables us to precisely understand the contribution of the predictions from the self-supervised model in solar nowcasting--if the predictions are useful, it will improve nowcasting results over using a persistence baseline's prediction. Moreover, this enables faster computation and cheaper memory overhead as the expensive CNN-LSTM model is not trained on each of the many solar sites.

We use the SVR auto-regressive model, discussed in \ref{sec:solar}, to forecast 15-minute ahead solar generation. 
We consider 4 different models to evaluate forecast at time $t$: \\
(1) Persistence model: this is again a simple past predicts the future baseline; \\
(3) CNN-LSTM-SVR: an SVR model using the predictions of CNN-LSTM, that is the forecasted channel values from past values $(C_{t-1}$,$C_{t-2}$,$C_{t-3})$ from the self-supervised CNN-LSTM model; \\
(3) SVR($C_{t-1}$): using the persistence model on satellite observations, $C_{t-1}$, as the forecast input to SVR instead of CNN-LSTM forecast, this should be an upper-bound on the error only if the CNN-LSTM model produces useful forecasts;\\
(4) SVR($C_t$): this is a lower-bound on the error that uses the ground-truth satellite observation at the \textit{future} instant and is not a feasible forecast as $C_t$ is unavailable ahead of time.

Note that SVR($C_t$) uses the current satellite observations to make predictions. This is a lower-bound on the error of the model given a particular site with some historical data when using SVR auto-regressive models. Estimates of satellite channel values, $\hat{C}_{t}$ from a model using previous instants observations, will be useful if they lead to an accuracy that is closer-to this performance of using the actual observations. Similarly, SVR($C_{t-1}$) corresponds to using persistence model predictions as the future satellite channel estimates, assuming no change in values, and will be an upper bound on the error. Model error should be below this value for the model to be useful for solar nowcasting.

\renewcommand{\arraystretch}{1.8}
\begin{figure*}[h]
\centering
\begin{tabular}{@{}c@{}c@{}}
\includegraphics[width=0.5\textwidth]{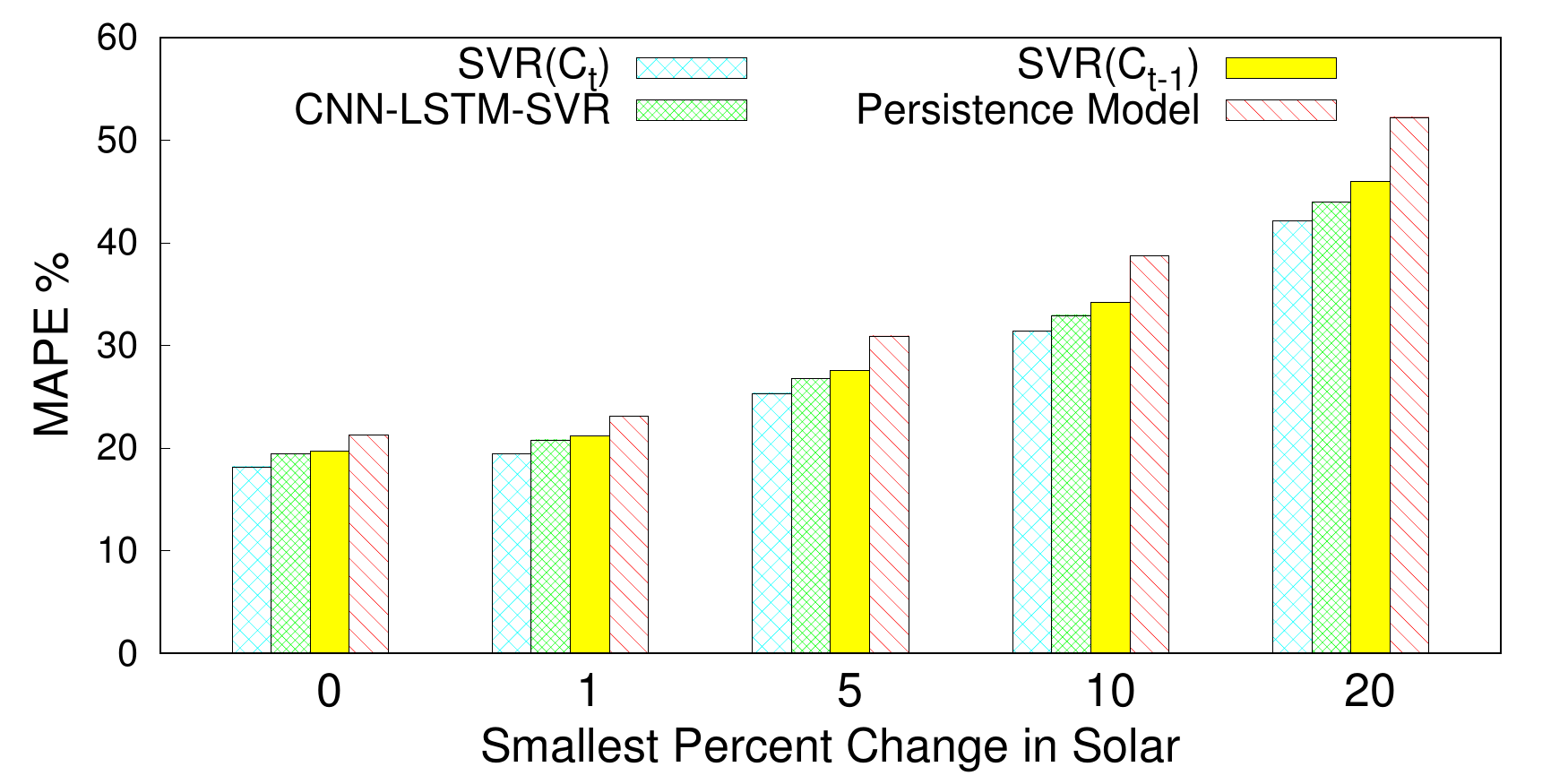} &
\includegraphics[width=0.5\textwidth]{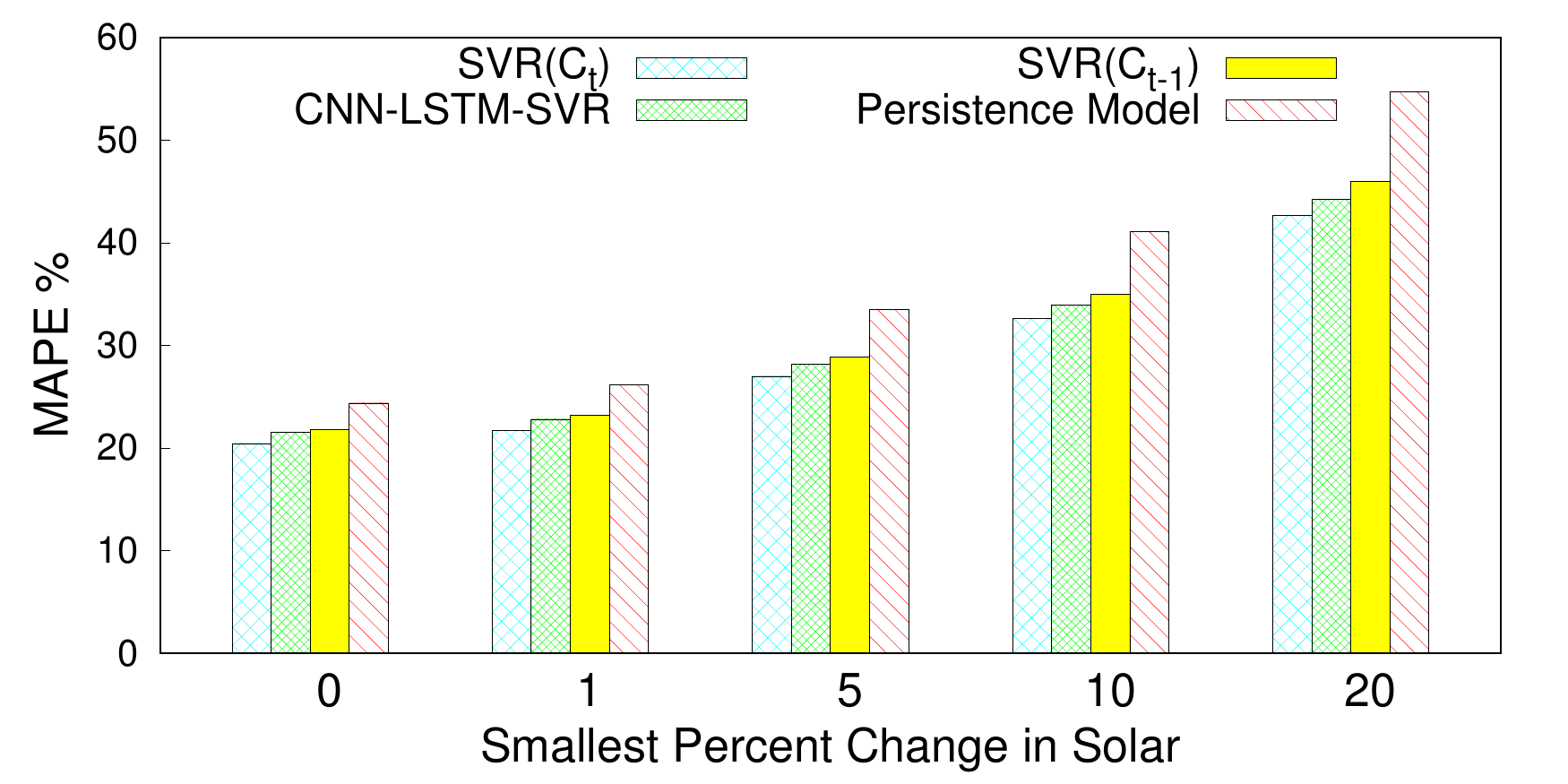} \\
\end{tabular}
\vspace{-0.3cm}
\caption{End to end solar forecasting on 10x10 km area, averaged over 25 solar sites over 15 mins. Performance for summer months (May-September) is shown on left and for the full year on the right. Using the predictions from the CNN-LSTM model leads to solar output forecasting with error close to that of using the current satellite observations. Compared to precision model forecasts, this approach is consistently better, specially at predicting when there will be large changes in solar ($\ge5\%$).}
\label{fig:endtoend}
\end{figure*}

Our results are shown in Figure~\ref{fig:endtoend}, and consider both scenarios of only summer months and the whole year. 
We include both scenarios, as typically forecasting is easier over summer months, due to largely sunny days, and more difficult over the rest of the year, due to phenomenon like rain, clouds, and snow.
The performance of forecasting solar using CNN-LSTM is close to using the ground truth channel values from the future in the model, an upper-bound, hence showing that the approach is useful and accurate for solar forecasting. 
We have further split the performance of these models into percent changes between successive solar generation values, as shown on the x-axis, where 0 means any change and includes all the values, whereas as 5\% means a change of at least 5\% in subsequent values and so on.
We can also compare the results in the left and right plots of Figure~\ref{fig:endtoend} in that they both show similar trends but only differ in the MAPE, which is higher for a full year and a little lower for only summer months.
Interestingly, we find that these models are not drastically worse when evaluating over non-summer months, which indicates that the models capture rich spatio-temporal phenomenon from the satellite data for accurate modeling.


\begin{figure}[t]
\begin{center}
\includegraphics[width=0.9\textwidth]{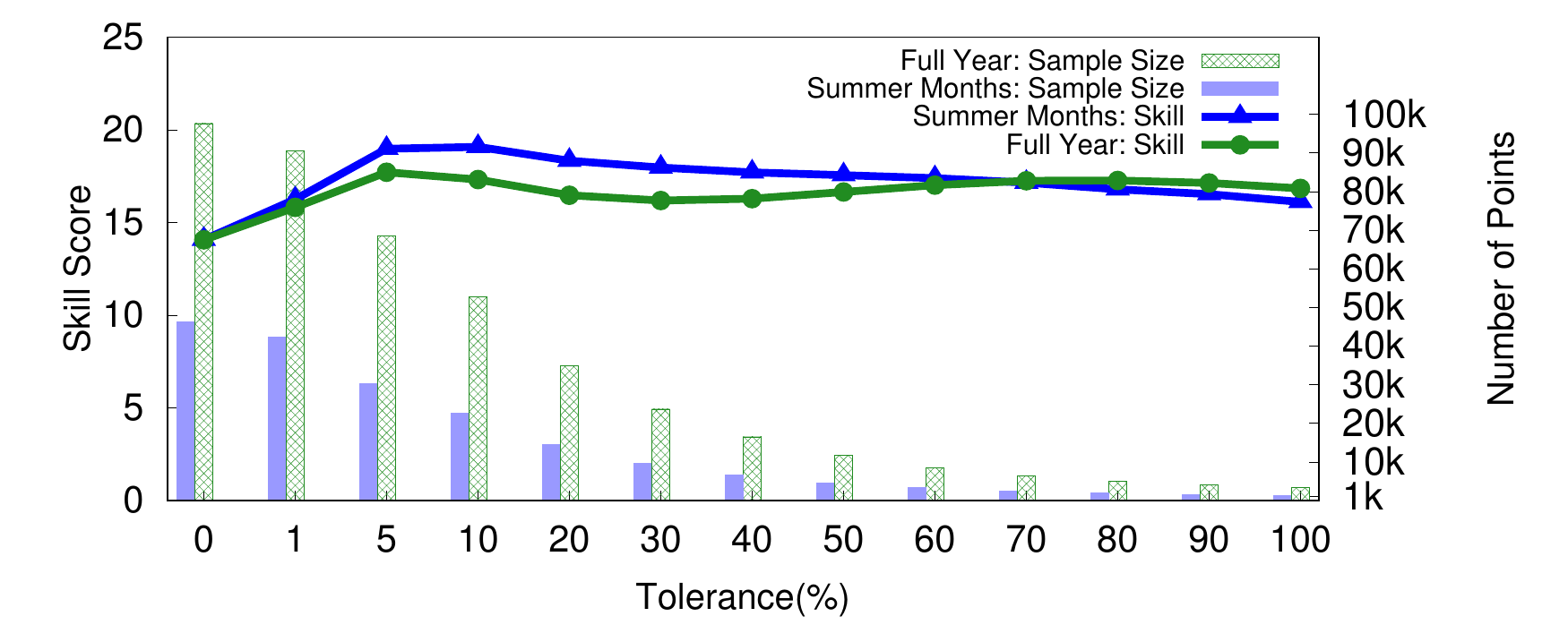}
\caption{Forecast skill score distribution over summer months and full year, along with the distribution of the number of instances at different tolerances.}
\label{avgss}
\end{center}
\vskip -0.1in
\end{figure}

\begin{figure}[t]
\begin{center}
\includegraphics[width=0.9\textwidth]{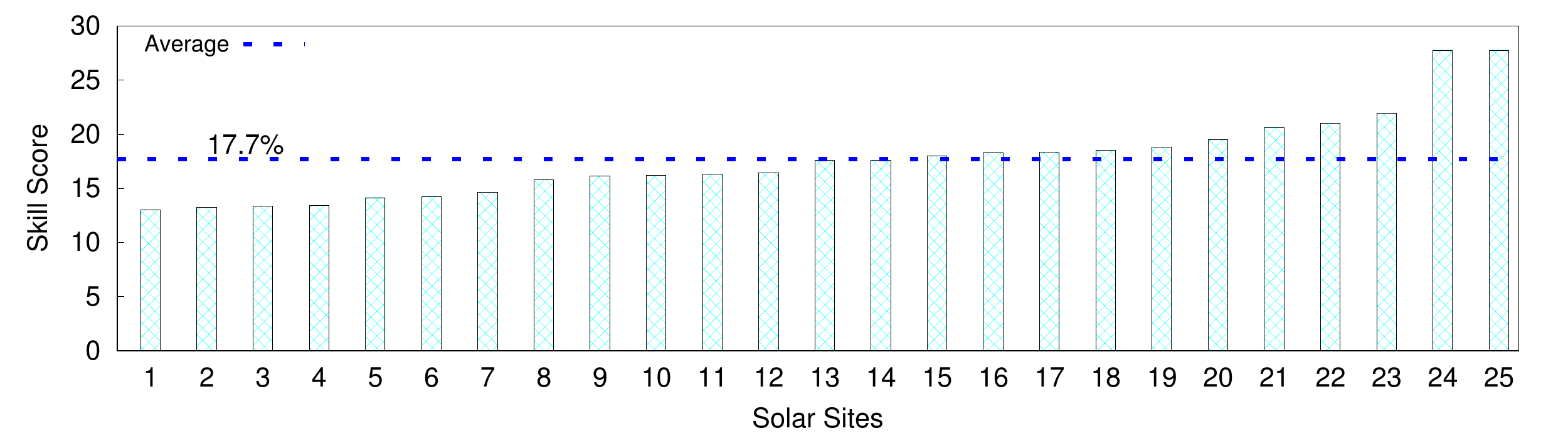}
\caption{Forecasting skill score of all the 25 solar sites at 5\% tolerance for the full year.}
\label{skillscorefullyr5}
\end{center}
\vskip -0.1in
\end{figure}

Skill score is a popular metric used to understand the performance of solar forecasting models. For solar nowcasting, the persistence model is the default baseline model that is used in prior work \citep{zhang2018deep,wang2019review,sun2019short,nie2020pv}.
Figure \ref{avgss} shows the average of the skill score across all the 25 sites at different tolerances.
In addition, we show the distribution of the number of data points available for evaluation at the various tolerances through a histogram.
We can see that the solar nowcasting model improves over the persistence baseline yielding average skill score in the range of 14-19\%.
We see that over full year, forecasting skill improves as we increase the tolerance of subsequent changes and then drops a bit at predicting very large changes of more than 5\%.
Interestingly, skill score is consistently high at predicting very large changes during summer months.

Figure \ref{skillscorefullyr5} then shows the distribution of forecast skill across the 25 solar sites. We can see that the skill varies widely across solar sites, from 14\%-27\% across the 25 sites. These variations are expected as different sites have different characteristics contributing to errors, including differences in installation capacities, shading from nearby buildings or trees, and widely different climates (snow vs sunny).
In particular, note that prior work on solar nowcasting using sky-camera imagery has found that state-of-the-art nowcasting models using deep learning have a skill in the range of 10\%-20\% \citep{nie2020pv,sun2019short}, as evaluated on only 1 or 2 solar sites (typical in sky-camera nowcasting research). The results in Figure \ref{skillscorefullyr5} show that our results are competitive or better, while being highly scalable and cost efficient as they don't require installation of specialized hardware at each solar site of interest.


%% file: related.tex
Forecasting solar PV output is akin to forecasting solar irradiance, since the former strongly correlates with the latter ~\citep{raza2016recent}.
Numerical Weather Predictions (NWP) algorithms \citep{10.1115/1.4042972,8245549,8447751,MATHIESEN2011967}, that mostly leverage physics-based modeling, are often used for solar irradiance forecasting.
These physics-based models are most appropriate for forecast horizons on the scale of hours to days, and not near-term forecasts on the scale of minutes to an hour \citep{hao2019novel,wang2019review}.  
Over long-term horizons, the complex and non-linear evolution of climate patterns can be difficult to model, requiring knowledge of climate processes and the history of many atmospheric events over time that can cause subtle changes. 

On the other hand, at shorter time scales of 5 to 60 minutes, machine learning approaches have the potential to implicitly model local changes directly from observational data \citep{wang2019review,rolnick2019tackling}.
While there is recent work on analyzing images from ground-based sky cameras \citep{zhang2018deep,zhao20193d,siddiqui2019deep,paletta2020convolutional} for near-term solar forecasting, it requires installing additional infrastructure at the site. 
Another alternative is based on estimating cloud motion vectors \citep{lorenz2004short, lorenz2012prediction, cros2014} from satellite images, however ML approaches that more directly model solar irradiance tend to perform better \citep{lago2018short,dsr2020}.  Our approach differs from recent approaches in solar nowcasting by forecasting solar irradiance values from multispectral satellite data using a combined CNN-LSTM, which can forecast changes in spatial features over time.  We then combine these solar irradiance forecasts with a model that predicts a site's solar output from solar irradiance. 

Our approach is a self-supervised approach, where we directly use abundant satellite data for modeling. Such methods have gained increasing popularity in computer vision recently \citep{jing2020self}.
While such methods have been widely successful, their applications in remote sensing have been limited and their applications to solar modeling have not been explored before.
\citet{jean2018tile2vec} similarly uses self-supervised learning over Landsat images, although their approach is designed for classifying geographical regions and not applicable here. 
\citet{vincenzi2021color} reconstructs visible bands from other bands, in a colorization task, to learn useful representations for land cover classification.
Related to our work, \citet{ayush2020geography} uses temporal information for constructing positive-negative pairs for classification of remote sensing data. Unlike this work, they ignore complex spatio-temporal dynamics and auto-regressive modeling, which are more crucial for forecasting.
Recently, parallel work \citep{ravuri2021skillful} utilized generative modeling on radar data for precipitation nowcasting using generative adversarial networks. Compared to this, we focus on a different application of solar nowcasting and are able to demonstrate utility of a simpler model for this application where directly predicting the next instant values is sufficient as they directly correlate with solar irradiance \citep{buildsys} -- hence do not necessitate requiring a complex discriminator for learning as in GAN models \citep{goodfellow2014generative}.
Finally, our work is also related to auto-regressive language models, that learn to predict the next word given previous words, which have been very successful for natural language processing \citep{radford2019language}.

%% file: conclusion.tex
Our work shows how to apply deep learning to multispectral satellite data to forecast short-term changes in solar output.  We propose deep auto-regressive models that combine CNN and LSTMs, trained in a self-supervised manner on abundant satellite data from GOES-R satellites. Such self-supervised training captures rich spatio-temporal dynamics that help improve solar nowcasting and is readily applicable to any solar site of interest, that is captured by the GOES-R satellites, without needing any specialized hardware like prior work. We evaluate our approach for different coverage areas and forecast horizons across a large number of 25 solar sites.  
Our results are promising and demonstrate that 15 minute forecasts using our approach have an error near that of a solar model using current weather and have forecast skill that is comparable with highly localized methods depending on installation of specialized sky cameras.  

While this is a promising first step, we believe there is much progress to be made in this area, as self-supervised learning is a promising approach for rapid progress in this domain due to abundant availability of rich satellite data.
The self-supervised learning approach itself can be improved by exploring longer contexts -- both in input and output, spatially and temporally -- through more sophisticated recent neural networks like Transformers \cite{vaswani2017attention} which can model longer range dependencies.
Finally, while solar nowcasting is the primary focus here, the self-supervised models presented here should be more generally useful for other applications such as detecting anomalies like wildfires, forecasting cloud cover, precipitation nowcasting, and more.